\documentclass[10pt, a4paper]{article}

\usepackage{lrec-coling2024} 
\usepackage{amsmath}
\usepackage{graphicx}
\usepackage{tabularx}
\usepackage{soul}
\usepackage{booktabs}
\usepackage{multirow}
\usepackage{xcolor}
\usepackage{color}
\usepackage{titlesec}
\usepackage{array}
\usepackage{float}
\usepackage{fontawesome5}
\usepackage{tikz}
\usetikzlibrary{positioning,shapes}

\newcommand*\circled[1]{\tikz[baseline=(char.base)]{
            \node[shape=circle,draw,inner sep=.6pt] (char) {#1};}}
\newcommand{\eqcontrib}{\hspace{.3em}\raisebox{.07em}{\resizebox{1em}{!}{\circled{\tiny\faEquals}}}}

\newcommand{\dataset}{\textbf{TGeGUM}}
\newcommand\blfootnote[1]{%
  \begingroup
  \renewcommand\thefootnote{}\footnote{#1}%
  \addtocounter{footnote}{-1}%
  \endgroup
}
\usepackage{xcolor}
\usepackage{hyperref}
 \definecolor{darkblue}{rgb}{0, 0, 0.5}
  \hypersetup{colorlinks=true, citecolor=darkblue, linkcolor=darkblue, urlcolor=darkblue}

\usepackage{xstring}
\usepackage[noabbrev,capitalize,nameinlink]{cleveref}

\title{{A}r{GUM}ent: Can Humans identify Genres and Topics?}
\title{Humans Versus Language Models for Genre and Topic Classification}
\title{Can Humans Identify Genres and Topics?}
\title{Can Humans Identify Domains?}

\name{Maria Barrett\textsuperscript{\faCompass\eqcontrib{}} \hspace{1em} Max Müller-Eberstein\textsuperscript{\faCompass}\textsuperscript{\faRobot\eqcontrib{}}\hspace{1em} Elisa Bassignana\textsuperscript{\faCompass}\textsuperscript{\faRobot\eqcontrib{}}\\ {\bf \large Amalie Brogaard Pauli\textsuperscript{\faWater\eqcontrib{}}\hspace{1em} Mike Zhang\textsuperscript{\faCompass}\textsuperscript{\faTint}\textsuperscript{\faRobot\eqcontrib{}}\hspace{1em}\vspace{1em} Rob van der Goot\textsuperscript{\faCompass}\textsuperscript{\faRobot\eqcontrib{}}}}

\address{\textsuperscript{\faCompass}IT University of Copenhagen, \textsuperscript{\faWater}Aarhus University, \textsuperscript{\faTint}Aalborg University,  \textsuperscript{\faRobot}Pioneer Centre for AI\\
         \textsuperscript{\faCompass}\texttt{\{mamy, elba, robv\}@itu.dk}\\
         \textsuperscript{\faWater}\texttt{ampa@cs.au.dk}, \textsuperscript{\faTint}\texttt{jjz@cs.aau.dk}\\}

\abstract{
Textual \emph{domain} is a crucial property within the Natural Language Processing (NLP) community due to its effects on downstream model performance. The concept itself is, however, loosely defined and, in practice, refers to any non-typological property, such as genre, topic, medium or style of a document. We investigate the core notion of domains via human proficiency in identifying related intrinsic textual properties, specifically the concepts of genre (communicative purpose) and topic (subject matter). We publish our annotations in \dataset: A collection of 9.1k sentences from the GUM dataset~\cite{zeldes2017gum} with single sentence and larger context (i.e., prose) annotations for one of 11 genres (source type), and its topic/subtopic as per the Dewey Decimal library classification system ~\cite{dewey1979}, consisting of 10/100 hierarchical topics of increased granularity.
Each instance is annotated by three annotators, for a total of 32.7k annotations, allowing us to examine the level of human disagreement and the relative difficulty of each annotation task. 
With a Fleiss' kappa of at most 0.53 on the sentence level and 0.66 at the prose level, it is evident that despite the ubiquity of domains in NLP, there is little human consensus on how to define them. By training classifiers to perform the same task, we find that this uncertainty also extends to NLP models. 
 \\ \newline  }

\begin{document}

\maketitleabstract

\section{Introduction}
\blfootnote{\eqcontrib{} All authors contributed equally.
}

The concept of ``domain'' is ubiquitous in Natural Language Processing (NLP), as differences between ``sublanguages'' have strong effects on model transferability~\citep{kittredge1986analyzing}. This issue of domain divergence has prompted comprehensive surveys on how to best adapt language models (LMs) trained on one or more source domains to more specific targets~\cite{ramponi2020neural, kashyap2021domain, saunders2022domain}, and remains an open issue, even with LMs of increasing size~\cite{ling2023domain,singhal2023large,wu2023bloomberggpt}. Despite its importance, what constitutes a domain remains loosely defined, typically referring to any non-typological property that degrades model transferability. In practice, textual properties with the largest domain effects relate to a document's genre/medium/style \cite{mcclosky2010any,plank2011domain,muller-eberstein-etal-2021-universal}, topic \citep{lee2001genres,karouzos-etal-2021-udalm}, or mixtures thereof (\citealp{aharoni-goldberg-2020-unsupervised}). More broadly, domains can be viewed as a high-dimensional space with variation across the aforementioned properties, plus factors such as author personality, age, or gender~\cite{plank2011domain,plank2016non}.

\begin{figure}
    \centering
    \includegraphics[width=\linewidth]{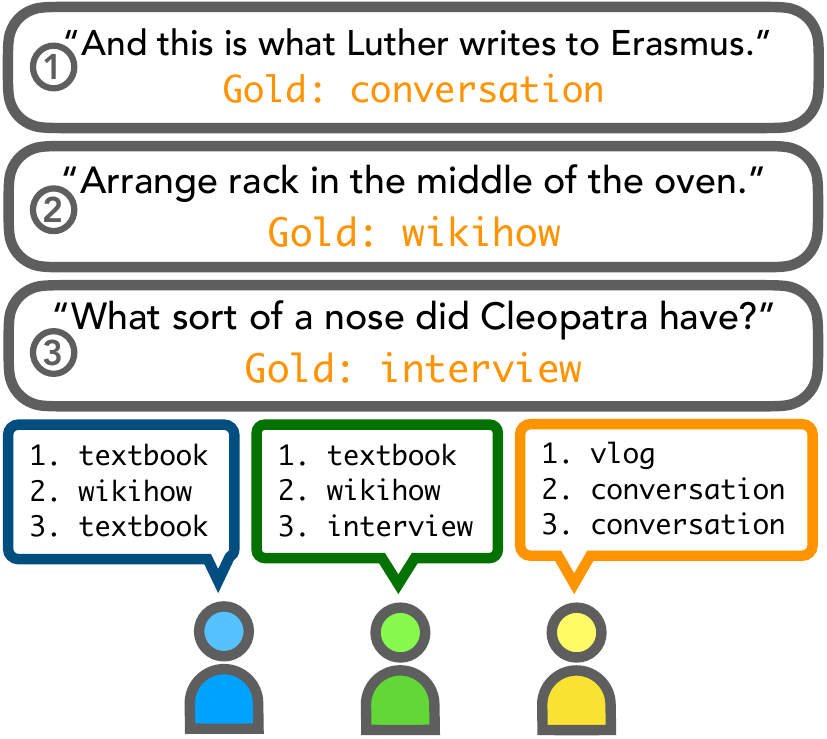}
    \caption{Graphical illustration of our triple-annotation setup with gold genre labels.}
    \label{fig:fig1}
\end{figure}

We attempt to gain a better understanding of the foundational concept of domain, by taking a step back from modeling this phenomenon, and instead investigating whether humans themselves can distinguish between different instantiations of domain-related properties of textual data. In linguistics literature, these properties are separated into register, style and genre \citep{biber1988variation,BiberDouglas2009RGaS,biber2019register}, of which we choose to focus on \textit{genre}, as it distinguishes itself from register and style by remaining consistent across complete texts. In addition, we examine the orthogonal factor of \textit{topic}, i.e., the subject matter of a text, which can be expressed independently of genre~\cite{kessler-etal-1997-automatic, lee2002text, stein2006distinguishing, webber-2009-genre}. We operationalize these two factors analogously to~\citet{van-der-wees-etal-2015-whats} as genre stemming from different source types with distinct communicative styles, and topic being the principal subject matter of a given text.

More formally, our main research question is: \emph{To what extent can humans detect genres and topics from text alone, and how does this align with machines?} We investigate the human proficiency in detecting these intrinsic properties by turning our attention to the Georgetown University Multilayer Corpus (GUM;~\citealp{zeldes2017gum}),\footnote{\url{https://gucorpling.org/gum/}} a large-scale multi-layer corpus consisting of texts from 11 different source types (henceforth \textit{genre}). These act as gold annotations against which we compare the manual genre labels provided by 12 human annotators for the entirety of the corpus (\cref{fig:fig1}). In addition, the annotators supply a new annotation layer regarding the texts' subject matter (henceforth \emph{topic}). As no gold labels are available for topic, they are annotated according to Dewey Decimal Classification (DDC;~\citealp{dewey1979}), a library classification system that allows new books to be added to a collection based on the subject matter. The DDC consists of 10 topics, 100 fine-grained topics, and 1,000 even finer-grained topics, of which we investigate the former two in detail and provide a preliminary study on the latter.

To understand the importance of context, we have annotators label genre and topic at both the sentence and prose level (defined as sequences of five sentences), and compare annotator agreement. Due to the subjective uncertainty associated with these types of characteristics, we gather three annotations per instance, measure their agreement, and release them in their unaggregated form as multi-annotations for future research.

Finally, we investigate the ability of machines to identify the same characteristics by training multiple ablations of genre and topic classifiers. Concretely, these experiments examine the difficulty of discerning each property, whether metadata or human notions of genre are more easily recoverable, as well as which level of context is most appropriate for the different ways in which the genre and topic label distributions can be represented.

Overall, this work is the first to explore the discernability of domain by both humans \textit{and} machines. In~\cref{sec:experiments}, we further discuss the implications of our findings, both with respect to domain-sensitive downstream applications, as well as for the NLP community's more general definition of domain. Our contributions thus include:

\vspace{.75em}
\begin{itemize}
    \itemsep0em
    \item \dataset{} (Topic-Genre GUM), a multi-layer extension of GUM, covering 9.1k sentences triple-annotated for a diverse set of 11 genres and 10/100 topics (\cref{sec:dataset}).\footnote{Data and code can be found at \url{bitbucket.org/robvanderg/humans-and-domains}.}
    \item An in-depth exploratory data analysis of the human annotations concerning annotator disagreement, uncertainty, and overall trends for domain characteristics across different context sizes (\cref{sec:eda}).
    \item A case study on the capability of NLP models to discern the human notions of genre and topic, as well as an analysis of which factors affect classification performance (\cref{sec:experiments}).
\end{itemize}

\section{Related Work}
\label{sec:relwork}
\paragraph{Domains} Initially coined as ``sublanguages'' \cite{kittredge1982sublanguage, kittredge1986analyzing}, domains have long been a topic of study in traditional linguistics and NLP~\cite{lee2002genres, lee2002text, stein2006distinguishing, eisenstein2014diffusion, van-der-wees-etal-2015-whats, plank2016non}. Some of the early work mentioning domains as textual categories include~\citet{sekine1997domain, ratnaparkhi1999learning}, which categorize texts into, e.g., ``general fiction'', ``romance \& love'', and ``press:reportage''. However, as also mentioned by~\citet{lee2002genres, lee2002text, plank2011domain, van-der-wees-etal-2015-whats}, the concept of domain is under-defined. \citet{plank2011domain} considers domains as a multi-dimensional space, spanning all kinds of variability between texts, such as genre, topic, style, medium, etc. In this work, we follow a definition of domains similar to~\citet{van-der-wees-etal-2015-whats}, focusing on two of the largest dimensions of variability: i.e., \textit{genres} (the communicative purpose and style) as well as \textit{topics} (the subject matter). The former is closely tied to the source of a text, such as academic papers versus fiction books, while the latter may include subjects such as sports, politics, and philosophy, which can occur in multiple genres.

\paragraph{Automatic Domain Detection} In NLP, automatic domain detection is essential for ensuring robust downstream performance, as it degrades with increasing levels of domain shift~\citep{ramponi2020neural}. Since this issue occurs independently of the application, domain classification has been explored in many contexts. Generally, the problem is either phrased in terms of a binary task, i.e., whether a target text matches the domain of the training data or not (e.g., \citealp{tan-etal-2019-domain,pokharel-agrawal-2023-estimating}), or a multi-label classification task, in which the exact domain is to be determined (e.g., \citealp{muller-eberstein-etal-2021-genre}). Here, we use the latter approach as it requires a more formalized operationalization of domain.

At a broader level, genre is frequently used as a proxy for domain, as it has lower internal variability than many more specific dimensions, including topic~\citep{kessler-etal-1997-automatic,webber-2009-genre}. Its automatic detection has been leveraged for selecting training data for transfer learning across a broad range of applications, such as classification ~\citep{ruder-plank-2017-learning,van-der-goot-etal-2021-effectiveness,gururangan-etal-2020-dont} and generative tasks~\citep{aharoni-goldberg-2020-unsupervised}. Beyond English, genre has further been shown to provide a cross-lingually consistent signal for enabling more robust transfer in syntactic parsing~\citep{muller-eberstein-etal-2021-genre}.

Topics provide a more granular differentiation between texts, also with close ties to domain. Automatically detecting topics has more immediate practical implications, as knowledge of the subject matter is critical for many downstream information extraction systems~\citep{crossNER,bassignana-plank-2022-crossre} and more datasets with topic annotations are available~\cite{sandhaus2008new, maas-etal-2011-learning, wang-manning-2012-baselines, zhang2015character}; however, these works typically contain source data from only a single corpus.

Going beyond prior work with limited sets of post-hoc topic labels for single-genre corpora, we build on the general-purpose DDC system~\citep{dewey1979} for libraries and apply its hierarchical set of 10/100 topics to a corpus containing data from 11 genres. By building on the existing annotations of the GUM dataset~\citep{zeldes2017gum}, we further enable research not only ascertaining to domain classification for its own sake, but also with applications to other downstream NLP tasks.

\paragraph{Multi-annotations} Given the subjective nature of domains and their associated properties of genre and topic, each text in our dataset is annotated multiple times and retains individual labels without aggregating them. This approach of \emph{multi-annotations}~\citep{plank-2022-problem} avoids obscuring human uncertainty in the annotation process and has benefits both for tasks with high variability, such as ours, as well as tasks for which a ground truth is typically assumed.

E.g., \citet{plank-etal-2014-linguistically} map part-of-speech (POS) tags from~\citet{gimpel-etal-2011-part} to the universal 12-tag set by~\citet{petrov-etal-2012-universal}, retaining five \emph{crowdsourced} POS labels per token.

For Relation Classification (RC),~\citet{dumitrache-etal-2018-crowdsourcing} obtained annotations for 975 sentences for medical RC, where each sentence is annotated by at least 15 annotators on average. 

For Natural Language Inference (NLI),~\citet{nie-etal-2020-learn} released ChaosNLI: A dataset with 4,645 examples and 100 annotations per example for some existing data points in the development set of SNLI~\cite{snli}, MNLI~\cite{mnli}, and Abductive NLI~\cite{Bhagavatula2020Abductive}. For a more in-depth overview of multi-annotation datasets, we refer to~\citet{uma2021learning}.

\begin{figure*}[t]
    \centering
    \includegraphics[width=\linewidth]{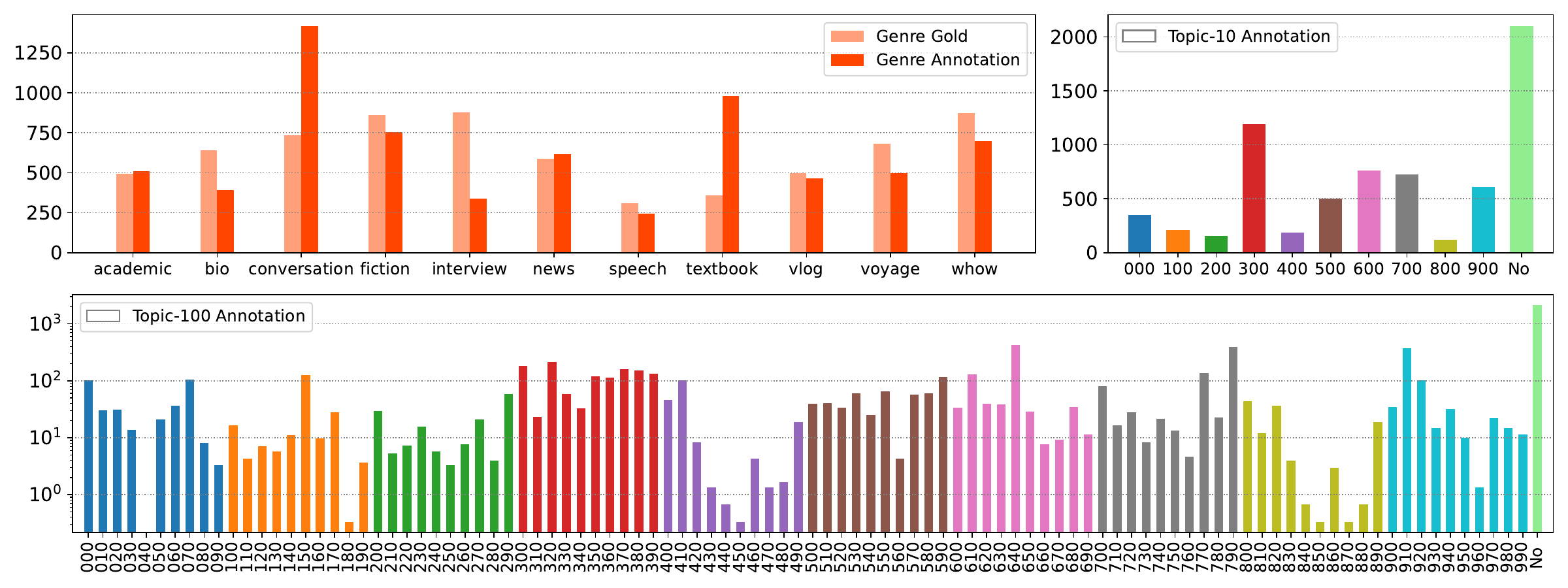}
    \caption{Frequency distributions of the labels in gold genre labels, annotations of genres, annotations of topic-10, and annotations of topic-100 (log scale) on sentence level. For the human annotations, the number is divided by three in order to align with the (unique) gold label. The mapping of topic-10 and topic-100 labels can be found in~\cref{sec:labels}. The tag ``No'' in the topic annotations refers to \textit{no-topic}.}
    \label{fig:dist-sent}
\end{figure*}

\section{The Dataset}\label{sec:dataset}

\subsection{Source Data}

The source dataset on top of which we build our domain-related annotations is the GUM corpus which in turn incorporates data from a wide variety of sources.
We use the portion of the GUM corpus released as part of the Universal Dependencies project (UD; ~\citealp{nivre2017ud}), i.e., excluding Reddit. 
Since a text's source is closely tied to its communicative purpose, we consider GUM's \textit{data source} metadata field of each instance as the gold genre label. For the topic, no equivalent gold label is discernible from the metadata.

The entire dataset is annotated both at sentence and prose level to investigate the importance of context for genre and topic annotation. For this purpose, we follow the gold sentence segmentation provided by GUM. We opted for these blocks instead of paragraphs, as the latter are not natural dividers for all text types and can have a high variety of conventions and functions across genres. To avoid the same annotator observing the same sentence individually as well as in prose, we shuffle the dataset such that annotations of a sentence with and without context are distributed across different annotators, while maintaining coverage of the full dataset.

\subsection{Annotation Procedure}
\label{sec:annotation-procedure}

Since there are no official descriptions of the genres in GUM, our annotation guidelines refer to the descriptions from the homepages of the websites of the source or the corresponding abstracts from Wikipedia.
For topic annotation, we follow the Dewey Decimal library classification system~\cite{dewey1979} consisting of 10/100/1,000 hierarchical topics of increased granularity. We consider the 10 high-level and the 100 mid-level classes for the coarse- and fine-grained topic annotations.
We constrain our guidelines such that topic-100 should always be a sub-type of topic-10. For example, if topic-100 is ``520 Astronomy'', then topic-10 should be ``500 Science''.
When none of the topic-100 labels fit the fine-grained topic of the instance, the annotators were allowed to leave the more specific topic blank, i.e., annotating topic-100 with the same label as topic-10.
In addition, we include the \textit{no-topic} label for when it is not possible to identify a specific topic from the provided text., such as for very short sentences, like ``Ok'' or ``I agree with that.''

We completed an initial annotation round of 20 instances with all annotators and authors of this paper to evaluate the guidelines and annotation setup. None of this data is included in the final dataset. We continued with groups of three annotators annotating different subsets of the data. 
After an introductory meeting, further unclarities were discussed asynchronously throughout the process. Annotators were asked to pose their questions in general terms and to not use direct examples as to not bias the other annotators on specific instances.
We did not conduct inter-annotator studies over the course of annotation and only had minor guideline revisions during the annotation process since we are mostly interested in human intuitions of genre and topic, and there are no gold labels for the topic task.

Annotators could indicate whether they were unsure about the annotation of a specific instance, and were also asked to provide notes/comments, if applicable. The annotation rate started at approximately 80--150 instances per hour. To ensure a similar amount of effort across annotators, we asked them to aim for approximately 150 instances per hour (also considering that annotation speed increases over time).

In total, we hired 12 annotators, who were paid 34,21 EUR per hour (before tax) for a total of 32 hours per person over a period of 4 weeks. The mean age was 27 ($\pm$2), and their highest completed education was equally split between a bachelor's and a master's degree. All rated their English skills as either C2/proficient or native. Seven annotators were reported to be female, three male, and two other/non-binary.

\subsection{Dataset Statistics}
\label{sec:data}
\cref{tab:data} shows the final dataset statistics of \dataset. The dataset includes around 9.1K sentences, and 1.8K prose, each of them annotated by three individual annotators for genre, coarse-grained topic, and fine-grained topic. 

\begin{table}
    \centering
    \resizebox{\linewidth}{!}{%
    \begin{tabular}{lrrrr}
    \toprule
                & \multicolumn{2}{c}{Instances}     & \multicolumn{2}{c}{Annotations}      \\
                & Sentence  &   Prose              &  Sentence  &   Prose                    \\
    \midrule
      Train     & 6,911     &   1,358              & 20,733     & 4,074                      \\
      Dev.      & 1,117     &   217                & 3,351      & 651                        \\
      Test      & 1,096     &   221                & 3,288      & 663                        \\
    \midrule
      Total      & 9,124     &   1,796              & 27,372     & 5,388                       \\
    \bottomrule
    \end{tabular}%
    }
    \caption{Dataset Statistics: Note that each instance has three associated annotations.}
    \label{tab:data}
\end{table}

In \cref{fig:dist-sent}, we report the sentence-level distribution of gold labels and human annotations, reporting the average number of annotations per label (total number of annotations divided by three annotators) to align with the singular gold genre metadata.
For topic-10 and topic-100 we only report the human annotations as no gold labels exist. 

Comparing gold and annotated genre labels, we observe a skew towards \textit{conversation} and \textit{textbook}. 
We hypothesize that this is due to the small amount of context an annotator receives. 
For example, the sentence ``Is that all that's left?'' with the gold genre label \textit{fiction} is annotated by all annotators as \textit{conversation}. Another example is the sentence ``Some of the greatest poetry has been born out of failure and the depths of adversity in the human experience.'' with gold label \textit{interview}. All annotators annotated this example as \textit{textbook}. 

For topic, we note that despite skewness, almost all 100 topics are used. The \textit{300 Social sciences} including, e.g., \textit{320 Political science} and \textit{370 Education}, stand out as being the most prevalent topics. The most frequent label, however, is \textit{no-topic}, indicating that it is challenging to identify a specific topic given only one sentence and that individual sentences can be associated with different topics, depending on the surrounding context.

The genre distribution at the prose level (~\cref{app:prose}) reveals a more accurate distribution for \textit{conversation}-like utterances; however, the general skew towards \textit{textbook} remains.
Concerning topic, the main contrast to the sentence-level distributions is the reduction of the \textit{no-topic} label, confirming that more context is crucial for this task.

\section{Exploratory Data Analysis}
\label{sec:eda}

\begin{table}
\resizebox{\columnwidth}{!}{%
\begin{tabular}{lrrr|r}
    \toprule
    & \multicolumn{3}{c|}{Kappa} & Maj. Acc. \\
    & Genre & Topic-10 & Topic-100 & Genre \\
    \midrule
Sentences & 0.5260& 0.5213 & 0.4239 & 67.68 \\
Prose & 0.6582 & 0.5238 & 0.3838 &  81.11 \\
\bottomrule
\end{tabular}}
\caption{Agreement scores across annotators, and accuracy of majority vote among annotators compared to gold genre labels.}
\label{tab:agreement}
\end{table}

\begin{figure}
\resizebox{\columnwidth}{!}{%
    \includegraphics{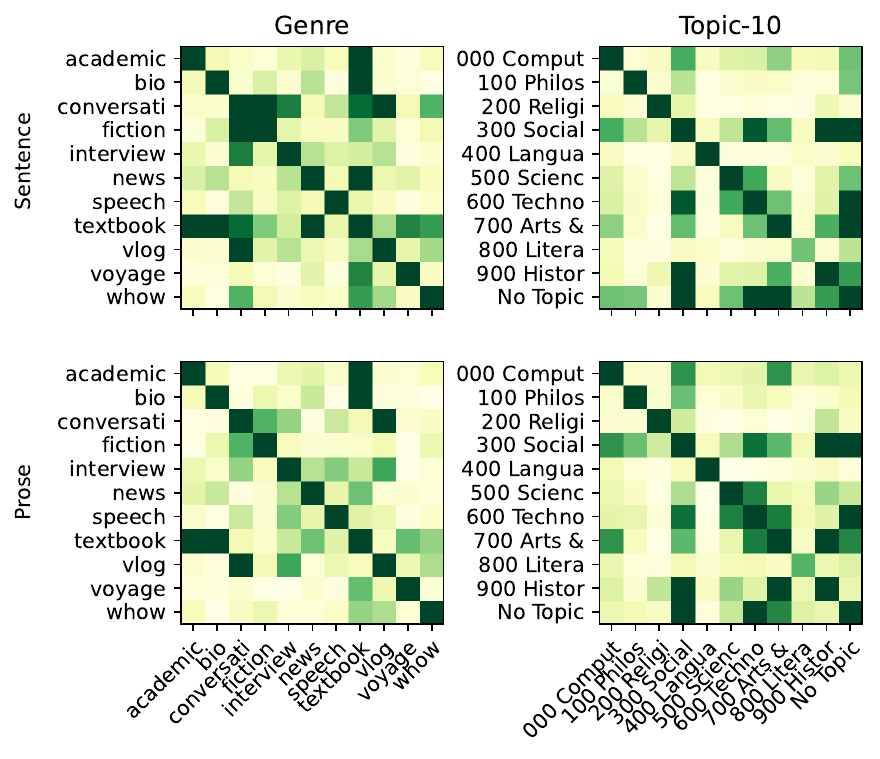}}
    \vspace{-1.5em} 
    \caption{Confusion matrix with all annotated pairs of labels for Genre and Topic-10 (across all annotators) in our training data: The darker the color, the higher the number of annotations for that label pair. The diagonal can be seen as agreement, whereas off-diagonal is a proxy for disagreement.}
    \label{fig:confusion}
\end{figure}

In addition to the previous aggregated overview, we are interested in exploring whether domain characteristics are recoverable by humans in a consistent manner. While we can compare human annotations to the original gold labels for genre, no equivalent is available for topic. Therefore, we place more emphasis on inter-annotator agreement, in the form of Fleiss' Kappa~\cite{fleiss1971measuring}, to measure intuitive alignment and ease of identification. \cref{tab:agreement} and~\cref{fig:confusion} shows this agreement across the different genres, topics and levels of available context.

\subsection{Human Genre Detection}
\label{sec:eda-genre}

\paragraph{Accuracy and Agreement} Considering that annotation guidelines were phrased to avoid any intentional alignment to an existing ground truth (i.e., annotators were unaware of the existence of gold genre labels), an accuracy of 67.68\% at the sentence level shows that genre is recoverable to a far higher degree than by random chance or by a majority baseline. This further increases to 81.11\% given more context at the prose level and is also reflected in the increase from moderate inter-annotator agreement (0.53) to substantial agreement (0.66).

The additional context appears to help differentiate genres that have more similarities to each other. This phenomenon is especially pronounced for spoken-language data, such as \textit{conversation}, \textit{interview} and \textit{vlog}, which differ with respect to genre-specific conventions such as who the speech is directed towards (i.e., bi-directional, interviewee, video viewer), or how formal the register is. Both properties are more easily discernible across multiple turns.

Nonetheless, even given more context, high amounts of confusion remain between certain genres such as non-fiction texts of the type \textit{academic}, \textit{biography}, and \textit{textbook}. These are intuitively similar to each other and may require even more context to distinguish. Generally, genres appear to lie on a more continuous spectrum that is difficult to discretize in conceptually similar cases.

\begin{figure}
    \includegraphics[width=\columnwidth]{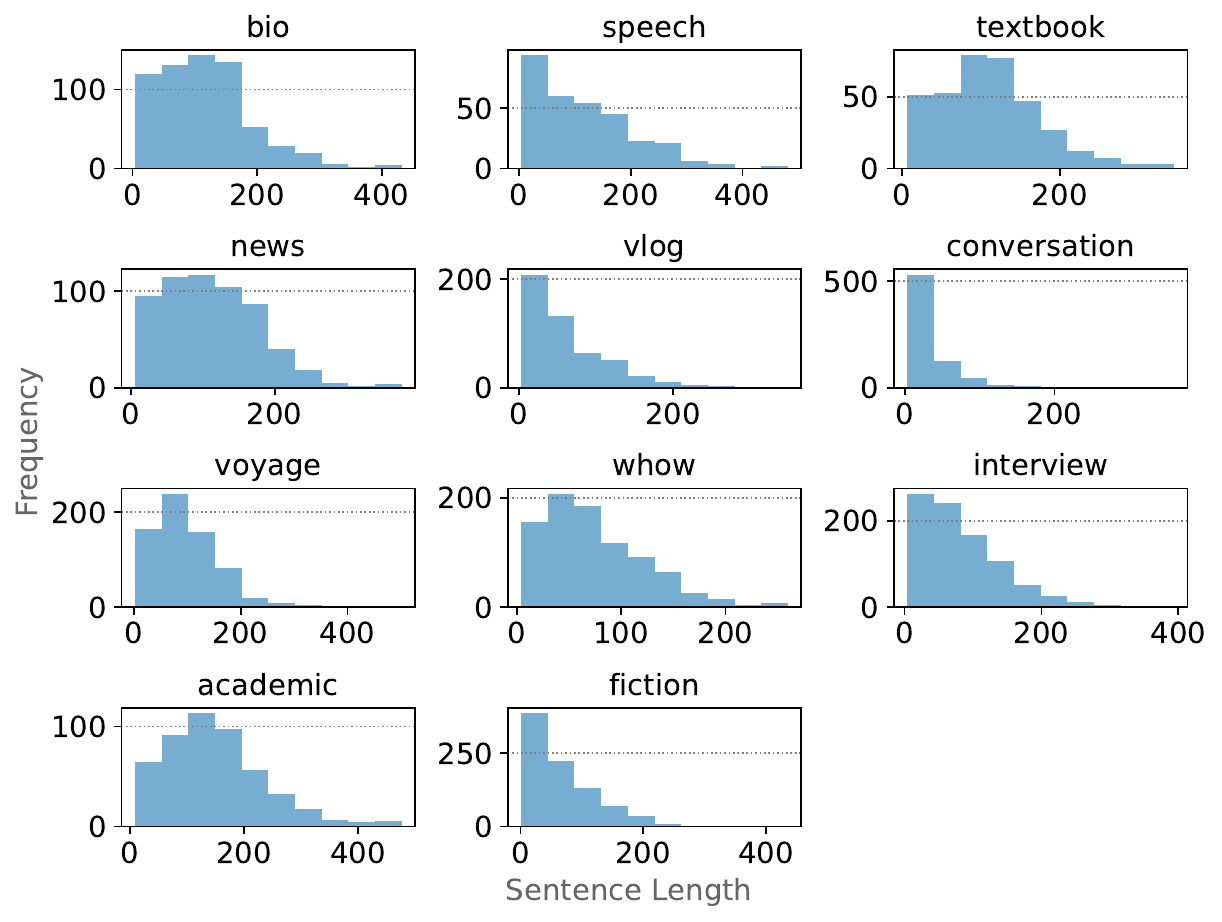}
    \vspace{-1.5em} 
    \caption{Frequency of sentence lengths, measured by the number of characters, per gold genre.}
    \label{fig:sent_len}
\end{figure}

\paragraph{Human Uncertainty} In case of uncertainty, annotators were encouraged to select a ``best guess'' label and to indicate uncertainty by ticking a checkbox. In addition to overall uncertainty, we also hypothesize that sentence length affects accuracy due to the amount of information available. To evaluate these two effects for genre detection, we measure the Pearson correlation between human accuracy concerning the gold label, with 1) sentence length, 2) the number of uncertainty flags (\cref{tab:correlations}). As expected, longer sentences are annotated correctly more often. \cref{fig:sent_len} further highlights how spoken-language genres have a strong skew towards shorter sentences, and for which annotators have the lowest agreements. Additionally, sentences marked as ``unsure'' align with gold labels less often, showing that annotators appear to have well-calibrated judgments of their own uncertainty, even for this relatively difficult task.

\subsection{Human Topic Detection}
\label{sec:eda-topic}

\paragraph{Agreement} In the absence of gold labels, inter-annotator agreement allows us to estimate the difficulty of discerning broader vs granular topics. For the 10 broader topics,~\cref{tab:agreement} shows a moderate agreement of 0.52 for both the sentence and prose levels. As expected with an order of magnitude more labels, Topic-100 sees a drop in agreement to 0.42 and an additional drop to 0.38 at the prose level. While this may seem counter-intuitive due to topic's higher specificity compared to genre,~\cref{fig:confusion} sheds some light on this peculiarity: In contrast to genre, topic has a \textit{no-topic} label (\cref{sec:annotation-procedure}), which, in turn, is used frequently by all annotators at the sentence level, due to the absence of any subject matter in many shorter utterances---especially in speech. Given the additional context, topic becomes more apparent, and agreement spreads toward more topics along the diagonal. As such, sentence-level agreement mainly hinges on \textit{no-topic}, while prose-level annotations agree more with respect to actual topics. This is less apparent for 10-topic kappa, for which this effect cancels out, but is more prevalent with 100 topics, where the shift away from \textit{no-topic} at the prose level comes with a much wider spread of topics, thereby reducing overall agreement, despite having a higher level of true topic annotations.

Overall, topics which were most consistently identified include \textit{social sciences}, \textit{arts \& recreation}, \textit{technology}, \textit{science} and \textit{history \& geography}. On the other hand, \textit{literature} was least consistently annotated and most frequently confused with the aforementioned topics, potentially due to its broader scope compared to the others.

\paragraph{1,000 Topics} After completing the full set of genre and topic-10/100 annotations with three annotators per instance, the remaining time of the annotators was spent on a preliminary study to label the most fine-grained categories of DDC. With 1,000 labels, this task is substantially more difficult. We obtained a total of 904 sentences and 172 prose sequences with three annotations each.\footnote{From 3,918 total annotations, we discarded instances with less than three completed annotations.} Measuring inter-annotator agreement at this level of granularity, we find a Fleiss' Kappa of 0.32 for sentences and 0.26 for prose. Although substantially lower than for coarser topic granularities as well as genre, this score still indicates above-random agreement among annotators. Similarly to the previous topic results, prose-level context allows humans to detect more actual topics than \textit{no-topic}, leading to lower overall agreement but a broader coverage of actual topics.

In general, despite the importance of topic to downstream applications (i.e., topic classification as a task in itself), there is no clear human consensus regarding discrete topic classification. Similarly to genre, topic appears to be a concept for which human intuition shares some agreement at a broader level, but is also spread along a continuum---especially as granularity increases.

\begin{table}
\centering
    \begin{tabular}{lrr}
    \toprule
        & Sent & Prose \\
    \midrule
    length vs unsure       & -0.1126$^*$   & -0.0474\\
    length vs correct      &  0.1267$^*$   & -0.0385\\
    unsure vs correct   & -0.2948$^*$   & -0.3411\\
    \bottomrule
    \end{tabular}
    \caption{Correlations across utterance length, correct predictions of human majority vote, and the number of unsure annotations. * indicates statistical significance for $p < 0.05$.}
    \label{tab:correlations}
\end{table}

\section{Modeling Domain}
\label{sec:experiments}

Following our examination of human notions of genre and topic, we investigate automatic methods' ability to model the same properties. Ablating across different setups for representing the multiple annotations per instance (\cref{sec:setup}), we train models to classify genre and topic at different levels of granularity (\cref{sec:results-classification}) and evaluate their ability to learn the underlying distribution (\cref{sec:results-distribution}). While pre-neural work typically performed document-level classification \citep{webber-2009-genre,petrenz2011genre}, contemporary trends have shifted towards the sentence-level \citep{aharoni-goldberg-2020-unsupervised,muller-eberstein-etal-2021-universal}. Leveraging our multi-level annotations, we investigate genre and topic classification at both the sentence and prose-level, mirroring our human annotation setup.

\subsection{Setup}
\label{sec:setup}

\begin{figure}
    \includegraphics[width=\linewidth]{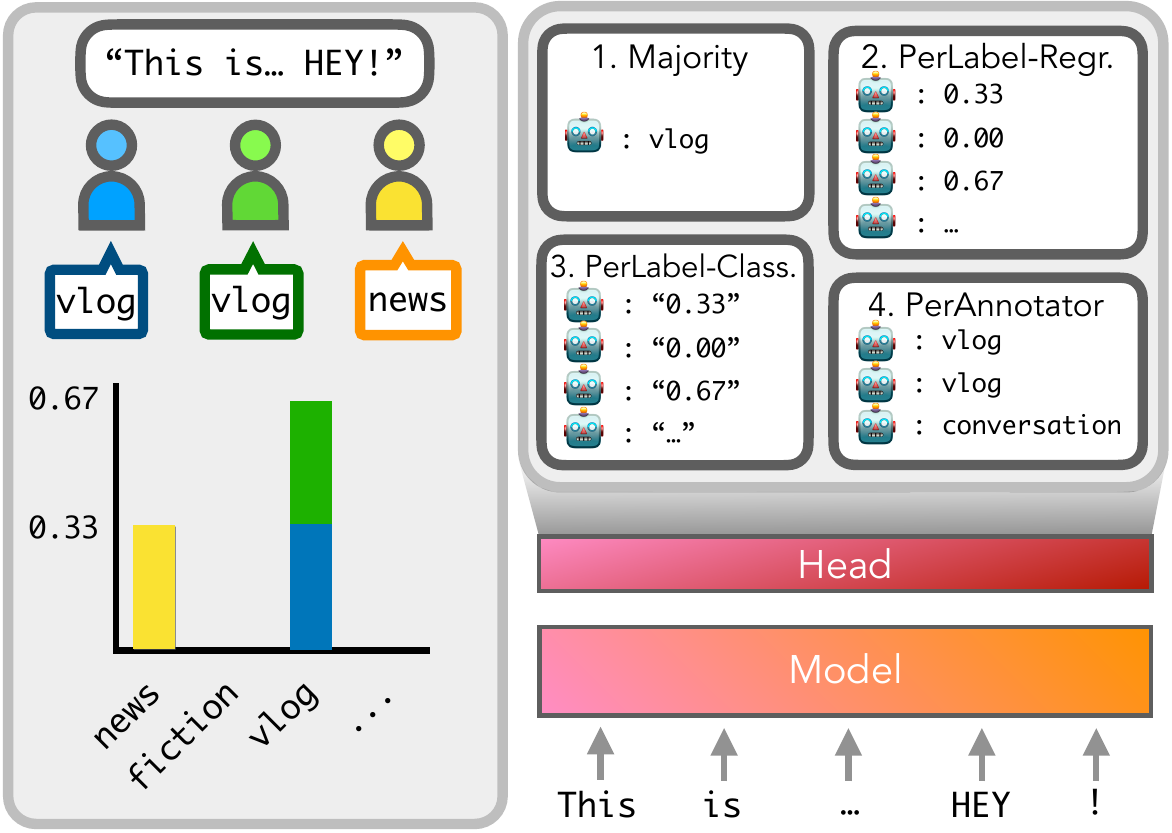}
    \caption{The target value each model variant is trained to predict: 1) Majority vote. 2) PerLabelRegr(ession) on label distributions. 3) PerLabel-Class(ification), on score bins per label. 4) PerAnnotator, three different annotations.}
    \label{fig:models}
\end{figure}

Most work on modeling multiple annotators is based on tasks consisting of only two or three labels, e.g., hate speech detection, or RTE~\cite{uma2021learning}. An exception is~\citet{kennedy2020constructing}, who use multiple classification heads to predict a score for a variety of aspects of hate speech, which are then used to predict a final floating point score for hate speech detection. Other related work predicts multiple task labels simultaneously ~\citep[e.g.,][]{demszky-etal-2020-goemotions,kiesel-etal-2023-semeval,piskorski-etal-2023-semeval}, however these are typically discrete and do not model annotator certainty.
We propose a variety of methods to model the distribution of the annotations (overview in ~\cref{fig:models}):

\paragraph{Majority} Discretizes the labels using a majority vote, and uses a single classification head to predict it. For the distribution similarity metric (see below), we assign a score of 1.0 to the chosen label.

\paragraph{PerLabel-Regression} Converts the human annotations to scores per label and then predicts these as a regression task. Each label has its own decoder head, trained using an MSE loss, and mapped to the [0;1] range afterwards.

\paragraph{PerLabel-Classification} Converts the human annotations into score bins and predicts them as four possible labels: ``0.0'', ``0.33'', ``0.66'', ``1.0''.

\paragraph{PerAnnotator} One decoder head modeling each annotator, that predicts their annotation as a discrete label. Afterwards, the three predictions are converted to a distribution.

\paragraph{} We evaluate these models using the standard accuracy over each singular predicted label (i.e., highest score or majority). In addition, we conduct a finer-grained evaluation that takes the multi-annotations into account. For this purpose, we propose a similarity metric for comparing the predicted and annotated label distribution per instance. Let $n$ be the number of label types, and $X$ and $Y$ are label distributions that sum to 1, with a score for each label. Then, the distributional similarity per instance can be computed as:

\[
distr\_sim = 1 - \frac{\sum_{l=0}^{n} | X_n - Y_n |}{2} \hspace{.3cm}\text{.}
\]

The resulting score between 0 and 1 represents the distributions' similarity. Note that we compare model predictions to the human annotations, which are not a gold standard; here, we aim to determine whether the human ability to discern these concepts is easy to model. 

We implement all our model variants in the MaChAmp~\cite{van-der-goot-etal-2021-massive} toolkit v0.4 using default parameters. MaChAmp is a toolkit focused on multi-task learning for NLP, and allowed us to implement all varieties of the tasks described earlier. Each way of phrasing the task is implemented on top of a single language model for fair comparison.
From an initial evaluation of the bert-large-cased~\cite{devlin-etal-2019-bert}, luke-large-lite~\cite{yamada-etal-2020-luke}, deberta-v3-large~\cite{he2021debertav3}, xlm-roberta-large~\cite{conneau-etal-2020-unsupervised} LMs on the gold genre labels, we identify that DeBERTa has the highest accuracy; hence we use it in the following experiments.

\subsection{Classification Results}
\label{sec:results-classification}
We examine which notion of domain is more learnable and distinguishable for a model; genre or topic? Since genre has associated ground truth labels, we additionally examine whether the human annotators' perception of genre or the ground truth genre is easier to learn. 

We establish a majority vote based on the human annotations; in case of a tie, the first element in the annotation list is chosen as the label, both for sentences and prose. This happens in $\sim$10\% of cases for genre and topic-10 (sentence and prose), and $\sim$20\% cases for topic-100.

~\cref{tab:dis_maj} shows accuracy and macro-F1 scores of the annotators' majority vote evaluated against the gold genre. As noted previously, more context (prose level) helps disambiguate the genre.

\begin{table}
\centering
\begin{tabular}{llll}
\toprule
 &          Accuracy &  Macro-F1 & $|N|$ \\ \midrule
Sentence   & 67.68 & 59.92 & 1,117 \\
Prose      & 81.11 & 74.75 & 217 \\ \bottomrule
\end{tabular}
\caption{Performance of annotators' majority vote compared with the gold genre (development set).}
\label{tab:dis_maj}
\end{table}

To evaluate how well a model can align with the human intuition of genres and topics, we fine-tune an LM on the majority labels of the annotators. 
We compare the performance on the gold genre labels (the only task for which we have gold labels) and compare the accuracy and macro-F1 scores (\cref{tab:dis_deberta}).
We notice the following:

\paragraph{Sentences} 1) Unsurprisingly, DeBERTa fine-tuned on the gold genre labels (gold\_genre) is better aligned with the ground truth genre than the human majority vote, i.e., \ 73.20 (\cref{tab:dis_deberta}) versus 67.68 (\cref{tab:dis_maj}) accuracy at the sentence level (note that other LMs performed worse).
2) In contrast, the fine-tuned DeBERTa model has higher accuracy when trained and tested on the human majority vote (maj\_gerne) than when using gold genre labels (gold\_genre), i.e., 75.88 versus 73.20, although macro-F1 is lower. This indicates that less common genre labels are easier to learn from gold labels, while more frequent genres are easier to learn based on human intuitions.
3) Despite topic-10 having fewer classes than genre, the notion of topic appears to be more difficult for a model to learn (lower F1). 4) The skew of the fine-grained topics (maj\_topic-100) and the difficulty of the long tail become apparent in the large divergence across the accuracy and macro-F1 score.

\paragraph{Prose} 5) In contrast to the sentence level, our fine-tuned DeBERTa model generalizes better to the gold genre labels (gold\_genre) than the human majority vote (maj\_genre). At this level of context, the majority vote topic is also harder for a model to learn than the majority vote genre.

\begin{table}
\begin{tabular}{llll}
\toprule                                                   &                          & Accuracy                 & Macro-F1  \\ \midrule
\parbox[t]{2mm}{\multirow{1}{*}{\rotatebox[origin=c]{90}{Sent.\hspace{.5em}}}} & gold\_genre & 73.20\small{$\pm$ 0.02} & 70.74\small{$\pm$ 0.02}\\
    & maj\_genre & 75.88\small{$\pm$ 0.01} & 67.04\small{$\pm$ 0.01}\\
    & maj\_topic-10 & 75.56\small{$\pm$ 0.02} & 60.54\small{$\pm$ 0.07}\\
    & maj\_topic-100 & 64.55\small{$\pm$ 0.00} & 18.43\small{$\pm$ 0.02}\\
    \midrule
\parbox[t]{2mm}{\multirow{1}{*}{\rotatebox[origin=c]{90}{Prose\hspace{.5em}}}} & gold\_genre & 89.49\small{$\pm$ 0.02} & 88.02\small{$\pm$ 0.03}\\
    & maj\_genre & 80.83\small{$\pm$ 0.01} & 74.97\small{$\pm$ 0.03}\\
    & maj\_topic-10 & 67.74\small{$\pm$ 0.01} & 50.35\small{$\pm$ 0.03}\\
    & maj\_topic-100 & 52.35\small{$\pm$ 0.01} & 16.04\small{$\pm$ 0.02}\\
\bottomrule
\end{tabular}
\caption{Accuracy and Macro-F1 on test split, for DeBERTa models fine-tuned and evaluated on gold genre, human majority vote for genre, and human majority vote for topic-10/100 (standard deviations across five seeds).}
\label{tab:dis_deberta}
\end{table}

\subsection{Distributional Results}
\label{sec:results-distribution}

\begin{figure}
    \includegraphics[width=\columnwidth]{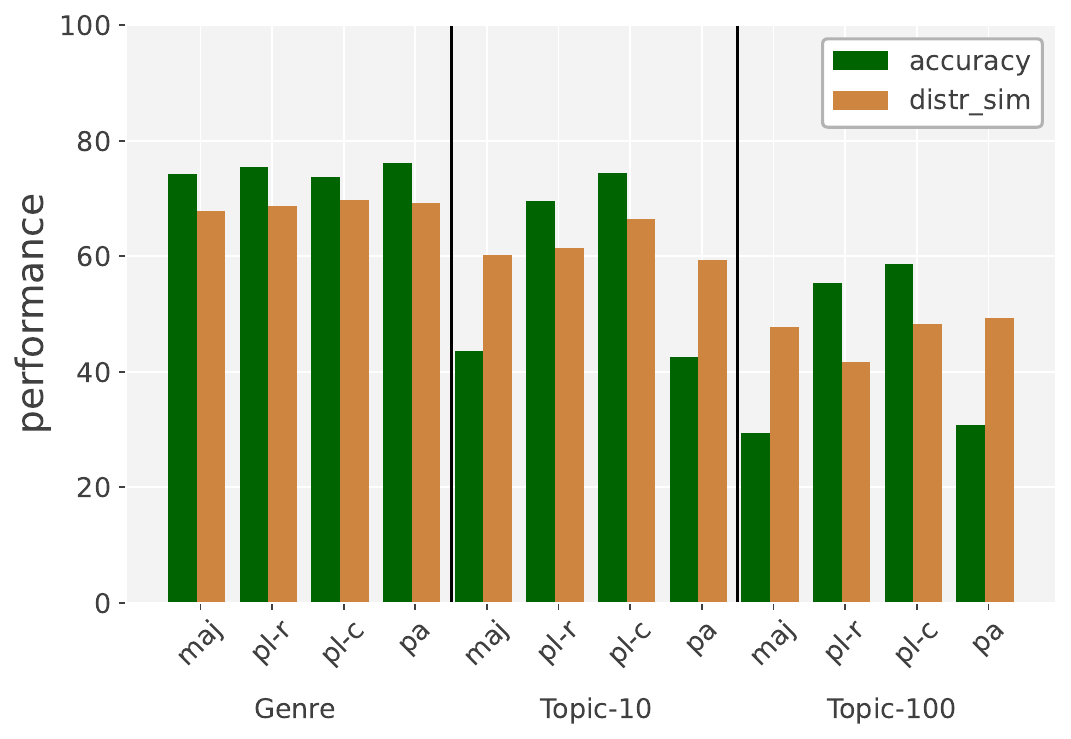}
    \vspace{-1em}
    \caption{Accuracy and distributional similarity on test split, for DeBERTa models trained on target labels based on Majority vote (maj), PerLabel-Regression/Classification (pl-r/c), PerAnnotator labels (pa); standard deviations across five seeds.}
    \label{fig:results-all}
\end{figure}

In~\cref{fig:results-all}, we report the results of the models trained on all instances (sentences and prose) with DeBERTaV3-large.\footnote{Training on sentences and prose separately leads to similar trends
(\cref{sec:sent-prose-results}).} 
The main trends show that the model performs better on the genre task. Unsurprisingly, for topics, the granularity of the labels impacts performance. 

By modeling the annotation distributions (i.e., PerLabel-Regression/Classification), we can outperform the majority vote model. However, distributional similarity decreases with increased label granularity (i.e., from topic-10 to topic-100), showing that it is difficult for models to calibrate to diverging human judgments. Interestingly, the per-label models achieve comparable or higher scores on the $distr\_sim$ metric, showing that the examined LMs model label distributions more easily than annotator behavior.

\section{Conclusion}

To examine the widely used but scarcely defined notion of \textit{domain}, this work provides the first investigation of human intuitions of this property in the form of \dataset{}: a collection of 9.1k sentences annotated with 11 genres and 10/100 topics by three annotators per instance, using an annotation procedure designed to capture human variability instead of forcing alignment (\cref{sec:dataset}).

Our exploratory analysis (\cref{sec:eda}) shows that despite the subjective nature of this task, as reflected in a Fleiss' Kappa of 0.53--0.66, humans can identify certain domain characteristics consistently from one sentence alone. Nonetheless, genres with a high similarity benefit substantially from added context. This is even more crucial for identifying topics, where we observe a shift from annotators not being able to discern any topic at all to being able to reach an above-random agreement, even when presented with 100 or 1,000 topics.

Finally, our experiments of modeling these domain characteristics automatically (\cref{sec:experiments}) show that genre is easier to model than topic. For both the agreements between human annotators, and the performance from the automatic model, we see that context is crucial for the genre classification task, but not for topic classification, where adding context even leads to decrease in scores if the label space is large.

Overall, this work highlights that despite the importance of ``domain'', there is little consensus regarding its definition, both in the NLP community as well as in our human annotations. Taking a closer look at what intuition predicted, further reveals that genres and topics are difficult to discretize completely, and that a continuous space of domain variability may be more suited for characterizing these phenomena.

\section{Ethics Statement}
Our approach to modeling human label variation is intrinsically linked to the larger issue of human social bias. As highlighted by \citet{plank-2022-problem}, significant social implications are tied to the study of label variation. In the context of our research, it is essential to acknowledge that variations in labeling might stem from societal biases and disparities. To address this, we recognize the necessity of addressing bias mitigation techniques as we aim to create more equitable and just models. However, we also contend that our focus on modeling generic subjects, such as genre and topic, may carry less severe implications compared to more subjective tasks like hate speech detection \cite{akhtar2021whose, davani2022dealing}. The differences in annotations within our work may primarily relate to two categories: ``Missing Information'' and ``Ambiguity''~\cite{sandri-etal-2023-dont}.

Another ethical facet we must address is the potential biases present in the classification system we use. In particular, the Dewey Decimal Classification System, which is the de-facto standard for libraries worldwide, has been found to exhibit prejudice~\cite{gooding2021dewey}. For example, the classification of information related to religion, specifically within class 200, demonstrates a clear skew, with a majority of subjects (six out of ten) reserved for Christianity-related topics. The remaining four slots are designated for other dominant religions, with an \textit{other} section meant to encompass all other belief systems. This reveals an inherent bias toward Christianity, which can affect the accessibility of non-dominant religions and belief systems. There are alternatives to knowledge organization systems like the Dewey Decimal Classification, as suggested by~\citet{franzen2022dewey}, to promote a more inclusive and equitable information landscape.

\section{Acknowledgments}
Many thanks to our annotators: Nina Sand Horup, Leonie Brockhaus, Birk Staanum, Constantin-Bogdan Craciun, Sofie Bengaard Pedersen, Yiping Duan, Axel Sorensen, Henriette Granhøj Dam, Trine Naja Eriksen, Cathrine Damgaard, and the other two anonymous annotators.
Maria Barrett is supported by a research grant (34437) from VILLUM FONDEN. Mike Zhang is supported by the Independent Research Fund Denmark (DFF) grant 9131-00019B. Elisa Bassignana and Max Müller-Eberstein are supported by the Independent Research Fund Denmark (DFF) Sapere Aude grant 9063-00077B.
Amalie Pauli is supported by the Danish Data Science Academy, which is funded by the Novo Nordisk Foundation (NNF21SA0069429) and VILLUM FONDEN (40516).

\section{Bibliographical References}\label{reference}

\bibliographystyle{lrec-coling2024-natbib}
\bibliography{papers}

 \clearpage

\section*{Appendix}
 \appendix
 
\section{Confusion Matrices Genre}
\begin{figure}
    \includegraphics[width=\columnwidth]{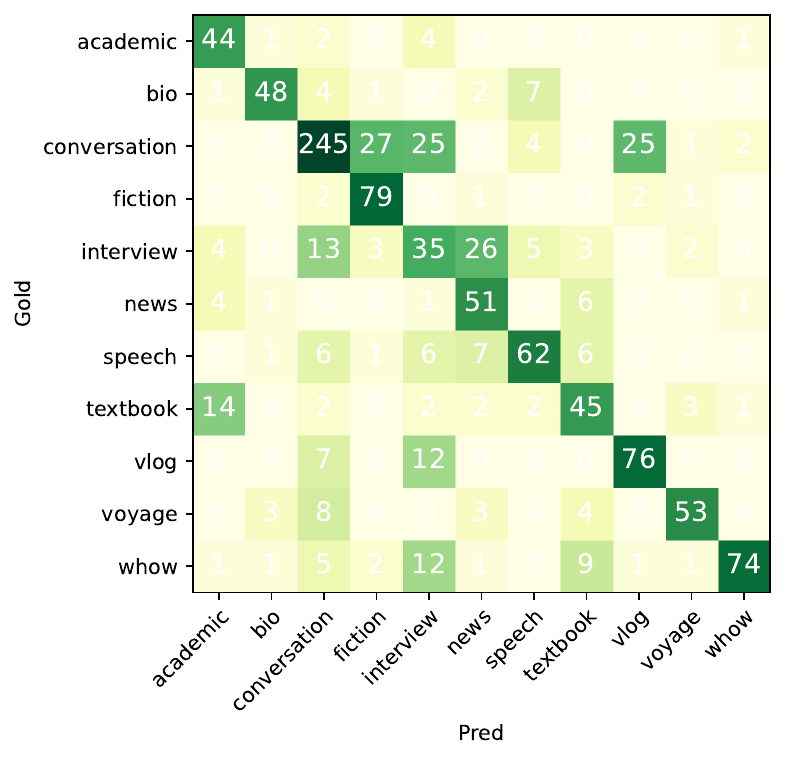}
    \caption{Confusion matrix on the sentence level, numbers are summed over all five random seeds.}
    \label{fig:genre-confusion-sent}
\end{figure}

\begin{figure}
    \includegraphics[width=\columnwidth]{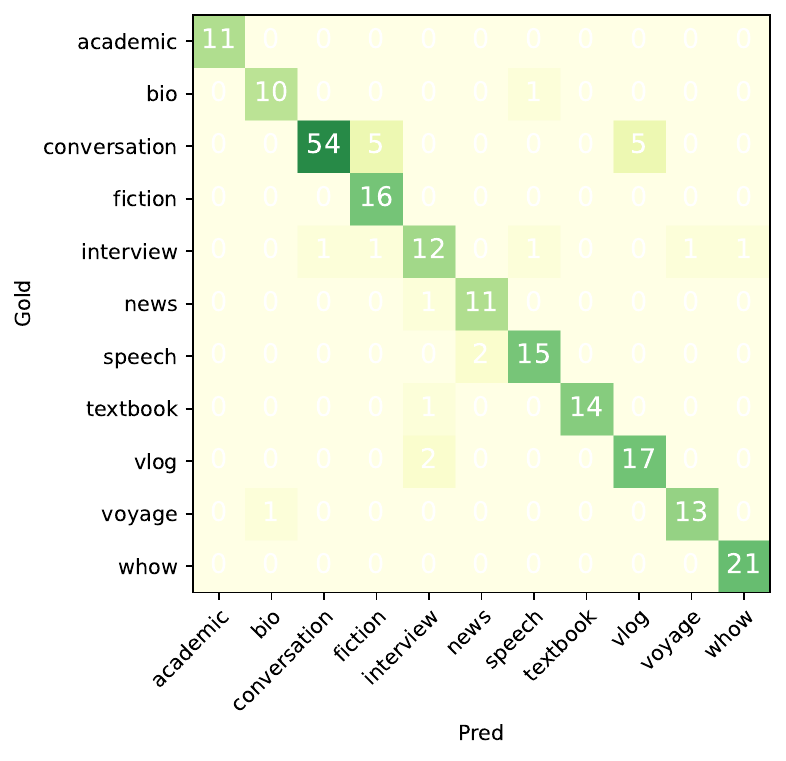}
    \caption{Confusion matrix on the prose level, numbers are summed over all five random seeds.}
    \label{fig:genre-confusion-para}
\end{figure}

\begin{figure}
    \includegraphics[width=\columnwidth]{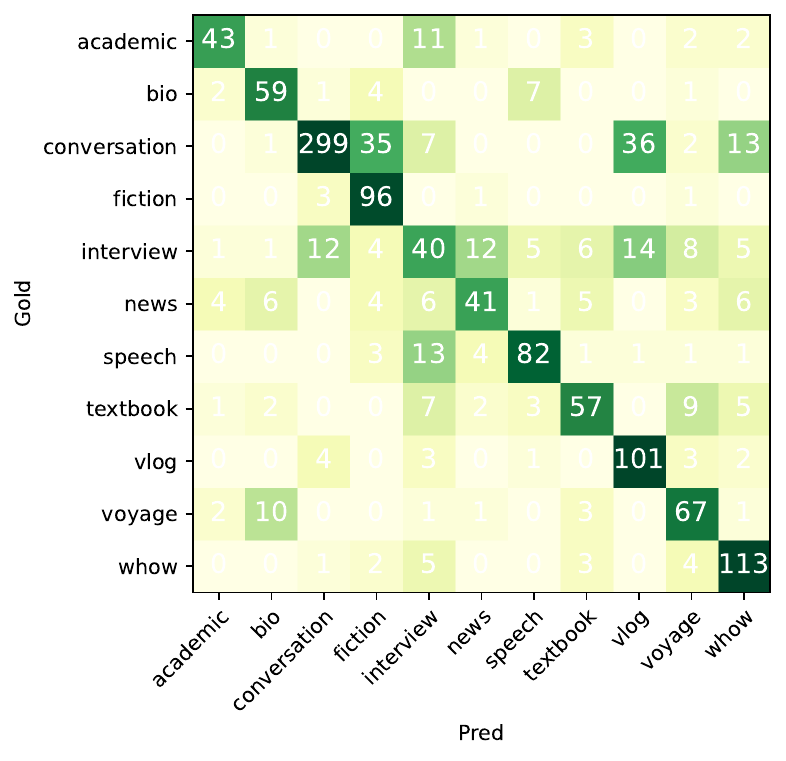}
    \caption{Confusion matrix on all data, numbers are summed over all five random seeds.}
    \label{fig:genre-confusion-all}
\end{figure}

In Figure~\ref{fig:genre-confusion-sent}-Figure~\ref{fig:genre-confusion-all} we plot the confusion matrices of our DeBERTa model trained on the gold genre labels. The conversation genre shows to be the most difficult label; it is commonly confused with fiction, interview and vlog; which also overlap in length (Section~\ref{sec:eda}).

\section{Sentence and Prose Results}
\label{sec:sent-prose-results}
\begin{figure}
    \includegraphics[width=\columnwidth]{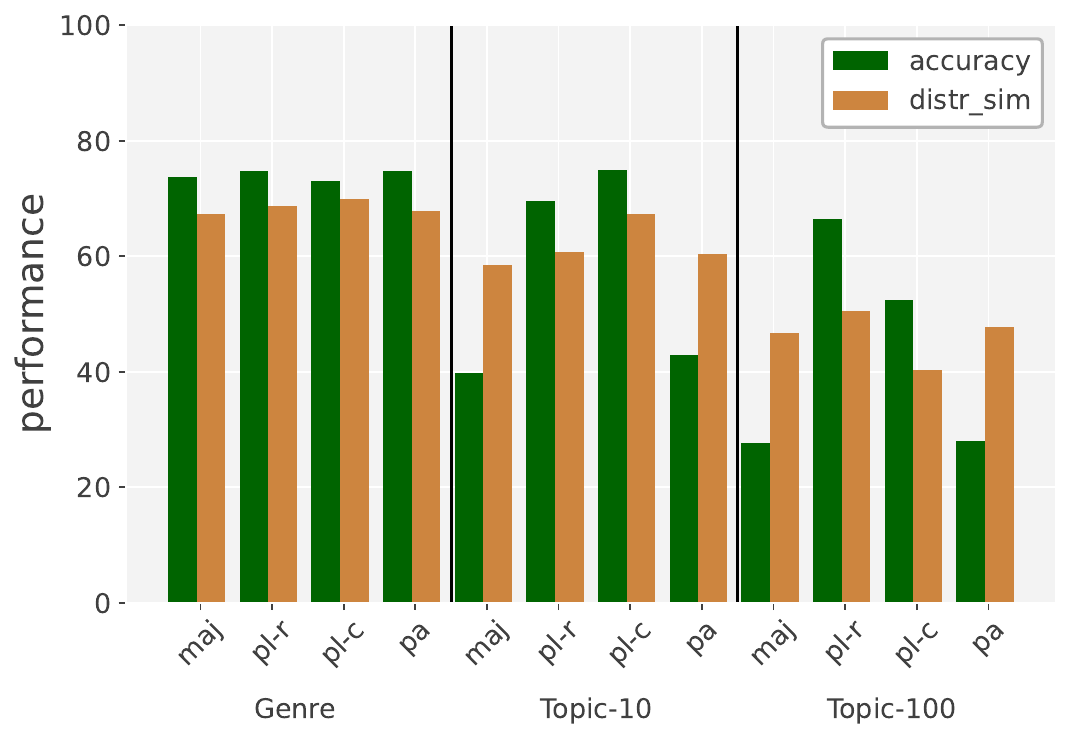}
    \caption{Results of our proposed models on the sentence level data.}
    \label{fig:distr-results-sent}
\end{figure}

\begin{figure}
    \includegraphics[width=\columnwidth]{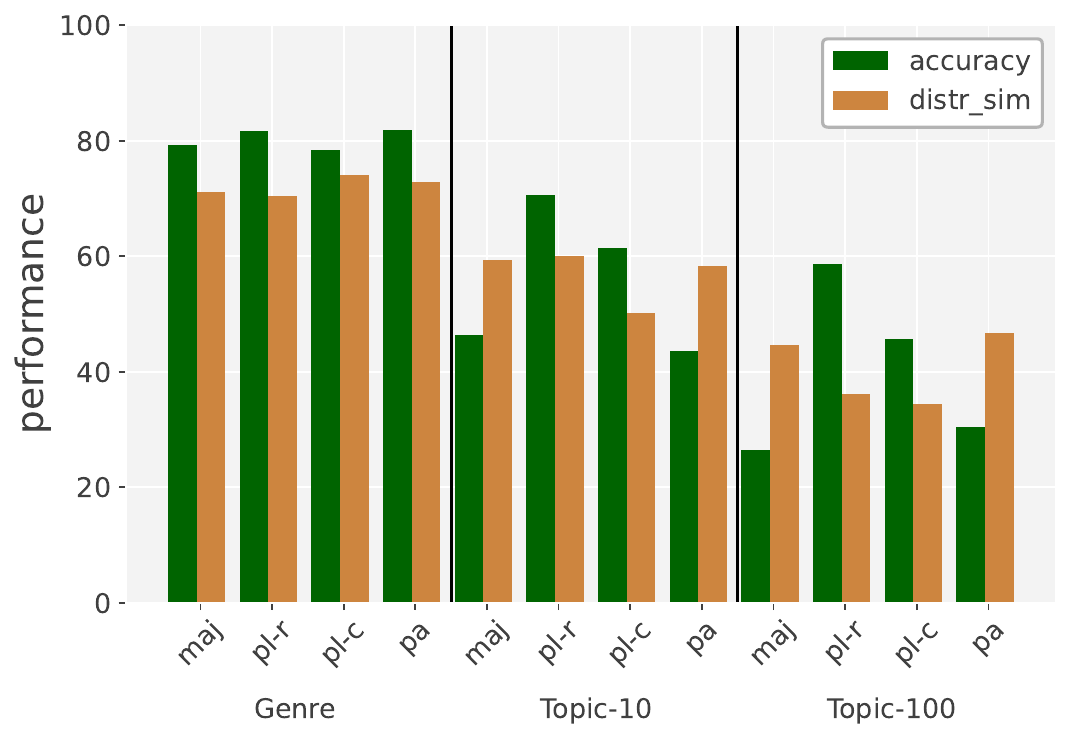}
    \caption{Results of our proposed models on the prose level data.}
    \label{fig:distr-results-para}
\end{figure}

In Figure~\ref{fig:distr-results-sent}  we show the results of our proposed models trained and evaluated only on the sentence level data. Figure~\ref{fig:distr-results-para} has the same evaluation on the prose level data. 

\section{Visualization of Embeddings}
\begin{figure}
    \centering
    \includegraphics[width=\columnwidth]{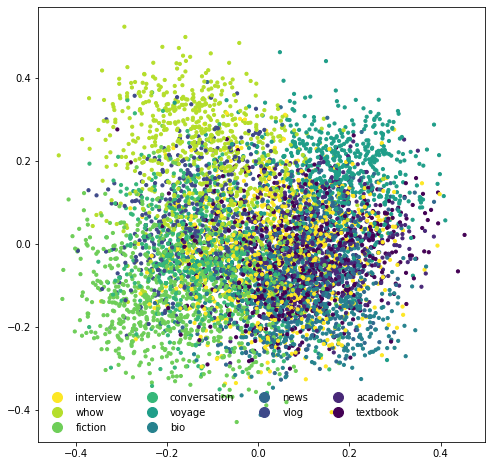}
    \caption{PCA plot of sentence embeddings with the gold genres.}
    \label{fig:pca_gold_genre}
\end{figure}

\begin{figure}
    \centering
    \includegraphics[width=\columnwidth]{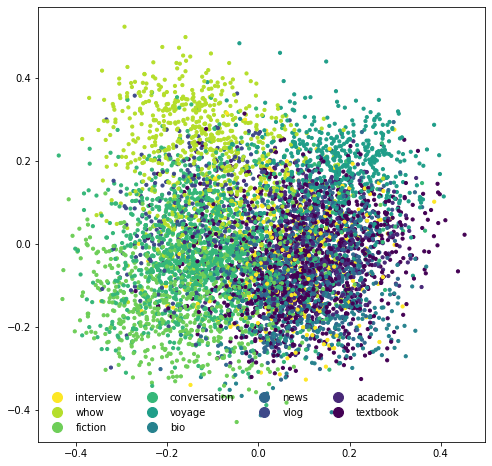}
    \caption{PCA plot of sentence embeddings with our annotation for genres, majority vote is used for each instance.}
    \label{fig:pca_maj_genre}
\end{figure}

\begin{figure}
    \centering
    \includegraphics[width=.95\columnwidth]{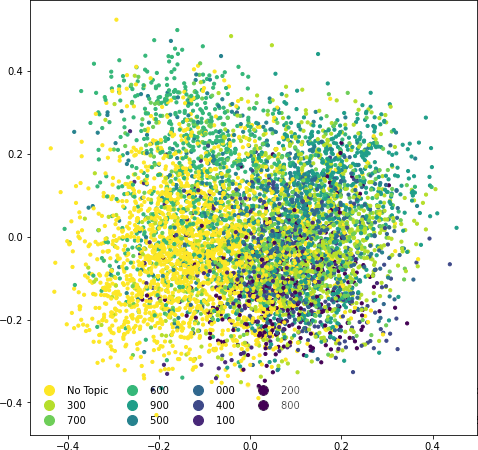}
    \caption{PCA plot of sentence embeddings with our annotation for coarse topics, majority vote is used for each instance.}
    \label{fig:pca_maj_topic1}
\end{figure}

We encode  sentences using Sentence-BERT~\cite{reimers-gurevych-2019-sentence}, apply a PCA-downprojection, and color each sentence according to gold genres, our majority-vote genre annotations, as well as majority-vote topic-10 annotations. The results are shown in Figures~\ref{fig:pca_gold_genre}--\ref{fig:pca_maj_topic1}.

\section{Prose-level Statistics}
\label{app:prose}
\begin{figure*}[t]
    \centering
    \includegraphics[width=\linewidth]{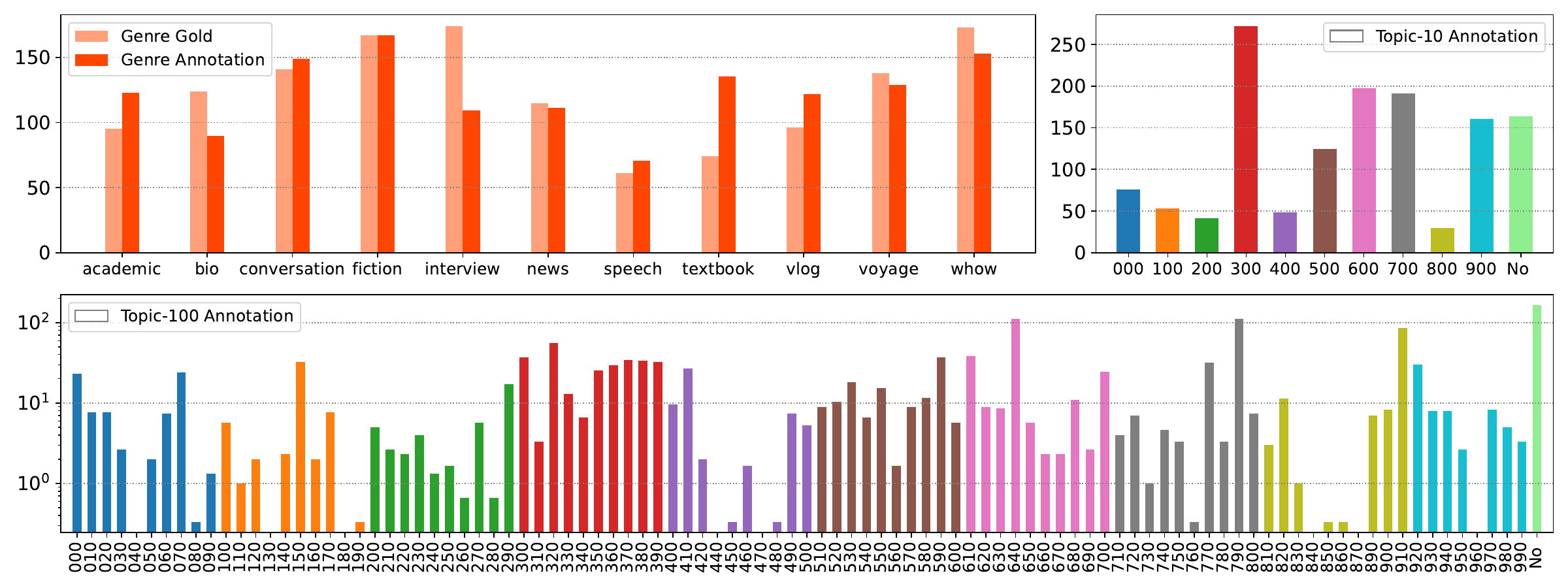}
    \caption{\textbf{Distribution of Labels (Prose).} Frequency distributions of the labels in gold genre labels, annotations of genres, annotations of topic-10, and annotations of topic-100 (log scale). For the annotations, the number is divided by three to get an average distribution. The mapping of topic-10 and topic-100 labels can be found in~\cref{sec:labels}. The tag ``No'' in the topic annotations means ``No topic''.}
    \label{fig:dist-prose}
\end{figure*}

Label statistics on the prose level are shown in Figure~\ref{fig:dist-prose}. While general trends, such as the majority genres and topics remain the same as on the sentence level, additional context spreads annotations more evenly, and allows for disambiguations such as for spoken data genres. This is also reflected in the higher alignment between gold and annotated genre labels---both in terms of number, but also in terms of accuracy (\cref{tab:agreement}). For topic, we further observe almost an order of magnitude fewer no-topic annotations, which are consequently distributed across the spectrum of actual topics.

\section{Annotator Comments}

Annotators were provided with a free-form field to provide optional comments regarding each annotation. Of the final dataset, 3.9\% of annotations have an annotator comment attached, with a median length of 38 characters. They primarily contain explanations of annotations which were marked with high annotator uncertainty.

\section{Guidelines}
\section*{Goal/Task}
In this annotation project, we are interested in knowing what the topic and genre is of a sentence and whether we humans can identify these. For Topics, we make use of the Dewey Decimal Classification (DDC) system. For genres, we make use of the genres provided in the Georgetown University Multilayer Corpus (GUM) corpus. The goal is to put the sentence/paragraph at hand into the most probable class (determined by you).

    \textbf{Genre} has a \textit{one-layer} annotation scheme, while \textbf{Topic} has a \textit{two-layer} annotation scheme, which we will refer to as L1 and L2. We want to annotate for all three. There is an option for "Not Sure" (abbreviated to "NS"). This is when you feel that the label for the sentence is not present in the options. In addition, feel free to add any notes for clarification (e.g., clarify your choice or something else).
    
\section*{Preliminaries}
Below we give an introduction to the topics and genre labels of this annotation project. It takes around 15-20 minutes to read. Note that you don't have to remember the label numbers. This introduction is to make you aware of the definition of the classes. All the labels are present in the annotation spreadsheet

\section*{Introduction Genres}
We make use of the text types (genres) in the GUM corpus. These genres do not have a specific number like the topics above. Therefore we simply enumerate them. The genres are the following:

\begin{itemize}
\item Academic
\item Bio
\item Conversation
\item Fiction
\item Interview
\item News
\item Speech
\item Textbook
\item Vlog
\item Voyage
\item Whow
\end{itemize}

\section*{Brief explanation of the genre classes}
\begin{itemize}
\item \textbf{Academic} (writing) is nonfiction writing adhering to academic standards and disciplines. It includes research reports, monographs, and undergraduate versions. It uses a formal style, references other academic work, and employs consistent rhetorical techniques to define scope, situate in research, and make new contributions.
\item A \textbf{bio}graphy is a detailed description of a person's life. It involves more than just basic facts like education, work, relationships, and death; it portrays a person's experience of these life events. Unlike a profile or curriculum vitae (résumé), a biography presents a subject's life story, highlighting various aspects of their life, including intimate details of experience, and may include an analysis of the subject's personality. Biographical works are usually non-fiction, but fiction can also be used to portray a person's life. One in-depth form of biographical coverage is called legacy writing. Works in diverse media, from literature to film, form the genre known as biography. An authorized biography is written with the permission, cooperation, and at times, participation of a subject or a subject's heirs. An autobiography is written by the person themselves, sometimes with the assistance of a collaborator or ghostwriter.
\item \textbf{Conversation}: naturally occurring spoken interaction. Represents a wide variety of people of different regional origins, ages, occupations, genders, and ethnic and social backgrounds. The predominant form of language use represented is face-to-face conversation, but lso documents many other ways that that people use language in their everyday lives: telephone conversations, card games, food preparation, on-the-job talk, classroom lectures, sermons, story-telling, town hall meetings, tour-guide spiels, and more.
Fiction refers to creative works, particularly narrative works, that depict imaginary individuals, events, or places. These portrayals deviate from history, fact, or plausibility. In our data, fiction pertains to written narratives like novels, novellas, and short stories. 
\item An \textbf{interview} is a structured conversation where one person asks questions and another person answers them. It can be a one-on-one conversation between an interviewer and an interviewee. The information shared during the interview can be used or shared with others.
\item \textbf{News} is information about current events, shared through various media like word of mouth, printing, broadcasting, electronic communication, and witness testimonies. It covers topics such as war, government, politics, education, health, environment, economy, business, fashion, entertainment, sports, and unusual events. Government announcements and technological advancements have accelerated news dissemination and influenced its content.
\item A (political) \textbf{speech} is a public address given by a political figure or a candidate for public office, usually with the aim of persuading or mobilizing an audience to support their ideas, policies, or campaigns. Political speeches are an essential tool for politicians to communicate their vision, articulate their positions, and connect with voters or constituents.
\item A \textbf{textbook} is a book containing a comprehensive compilation of content in a branch of study with the intention of explaining it. Textbooks are produced to meet the needs of educators, usually at educational institutions. Schoolbooks are textbooks and other books used in schools. Today, many textbooks are published in both print and digital formats. 
\item A \textbf{vlog}, also known as a video blog or video log, is a form of blog for which the medium is video. The dataset contains transcripts of the speech occurring in the video.
\item A travel/\textbf{voyage} guide is a wiki providing information for visitors or tourists about a particular place. It typically includes details about attractions, lodging, dining, transportation, and activities. It may also contain maps, historical facts, and cultural insights. Guide wikis cater to various travel preferences, such as adventure, relaxation, budget, or specific interests like LGBTQ+ travel or dietary needs.
\item A Wikihow how-to (\textbf{whow}) guide is an instructional document that offers step-by-step guidance on accomplishing a specific task or reaching a particular goal. It aims to assist individuals in learning and comprehending the process involved in successfully completing the task. These guides are typically written in a clear and concise manner, simplifying complex processes into manageable steps. They often include detailed explanations, diagrams, illustrations, or examples to enhance understanding. How-to guides cover various topics, such as technical tasks, practical skills, creative endeavors, troubleshooting, and more.
\end{itemize}

\section*{Introduction Topics}
\label{sec:labels}
The DDC system is a widely used library classification system developed by Melvil Dewey in the late 19th century. The DDC is based on the principle of dividing knowledge (in our case sentences) into ten main classes, each identified by a three-digit number; we only focus on the first two:

\begin{enumerate}
\item The ten main classes in the Dewey Decimal Classification system are as follows:

\begin{itemize}
\item 000 Computer science, information \& general works 
\item 100 Philosophy \& psychology 
\item 200 Religion 
\item 300 Social sciences
\item 400 Language 
\item 500 Science 
\item 600 Technology 
\item 700 Arts \& recreation 
\item 800 Literature 
\item 900 History \& geography
\end{itemize}
\textbf{These higher level classes belong to L1 in the annotation spreadsheet, and we added the NO-TOPIC label (see description below)}

\item 
Each main class is further divided into subclasses using additional digits (10s). For example, in the 500s (natural sciences and mathematics), you'll find 510 for mathematics, 520 for astronomy, 530 for physics, and so on. The system allows for more specific classification of books and materials based on their subject matter.

See the following page: \url{https://www.oclc.org/content/dam/oclc/dewey/ddc23-summaries.pdf}

This page separates the ten classes above into more finer-grained classes. There is not an explanation for each of them, but usually the name of the label encapsulates the subclass already. Note that the subclasses overwrite the main classes (so you can't pick 400 and 510, then you'd have to change 510 to 500).

\textbf{These subclasses belong to L2 in the annotation spreadsheet.}

Note that for each fine-grained class we deem the main number/code (e.g., 100, 200, 300) in L2 as the No-topic/Other category. The "Other" class can only be chosen in the fine-grained label classes (L2). Choosing this means that you believe that the current sentence belongs to a specific class. But the label is not present.
\end{enumerate}

The Dewey Decimal Classification system is used in many libraries around the world to organize their collections and make it easier for users to locate resources. It provides a systematic way of arranging materials and enables efficient browsing and retrieval of information based on subject areas.

\section*{Brief explanation of the topic classes (L1)}
\begin{itemize}
\item 000 Computer science, information \& general works is the most general class and is used for works not limited to any one specific discipline, e.g., encyclopedias, newspapers, general periodicals. This class is also used for certain specialized disciplines that deal with knowledge and information, e.g., computer science, library and information science, journalism. Each of the other main classes (100-900) comprises a major discipline or group of related disciplines. Note that in our experiments, we do not consider this a miscellaneous category, we have "No-topic" for this.
\item 100 Philosophy \& psychology covers philosophy, parapsychology and occultism, and psychology.
\item 200 Religion is devoted to religion.
\item 300 Social sciences covers the social sciences. Class 300 includes sociology, anthropology, statistics, political science, economics, law, public administration, social problems and services, education, commerce, communications, transportation, and customs.
\item 400 Language comprises language, linguistics, and specific languages. Literature, which is arranged by language, is found in 800.
\item 500 Science is devoted to the natural sciences and mathematics.
\item 600 Technology is technology.
\item 700 Arts \& recreation covers the arts: art in general, fine and decorative arts, music, and the performing arts. Recreation, including sports and games, is also classed in 700.
\item 800 Literature covers literature, and includes rhetoric, prose, poetry, drama, etc. Folk literature is classed with customs in 300.
\item 900 History \& geography is devoted primarily to history and geography. A history of a specific subject is classed with the subject.
\item No topic: For cases where the topic can not be determined, or even guessed. For example for utterances that contain no natural language or do not have enough context.
\end{itemize}

\section*{FAQ}
\begin{itemize}
\item Should the colors of L1 and L2 in the annotation spreadsheet match?

Yes, apart from that the colours should match, the first number of the class to which the sentence belongs should also match. 

For example, a sentence that belongs to Arts (700), is restricted to anything in the 700 class, e.g., a painting (750).

\item If a sentence has a clear topic in general, but the L2 category does not match, how do we annotate?

The fine-grained (L2) topics have the priority, and since they have to match you adjust the main topic accordingly.

\item Does my choice of Topic depend on the Genre or vice versa?

No, by default, annotating for genre and topic should be a separate task and should not influence each other.

\item How do we distinguish between something that is in the No-topic (or Others) class and NS ("not sure")?

Use the “others” category when you believe the current instance to belong to a class which is not in the listed ones. Mark your choice with “NS” when you have a guess, but you are not confident about it (e.g., because the instance is very short, or you are not familiar with the genre/topic)

If you are able to find L1, but none of the labels fit for the sentence in L2, you should choose "Other" (e.g, 000, 100, 200, etc.) in the same colour (class) of L2. The "Other" class can only be chosen in the fine-grained label classes (L2). Choosing this means that you believe that the current sentence belongs to a specific class. But the label is not present. Otherwise, mark your best guess with “NS”.

\item Is it better to label a sentence as "NO-TOPIC" if there is not a clear label associated with it or are we encouraged to take a guess?

You are encouraged to take a guess. However, for cases where you have no preference for any of the labels (i.e. a wild guess), label it as NO-TOPIC.

\item There is already another "Other" class in Religion/Language (e.g., 290 Other religion).

Good catch, imagine this situation. Let's say the sentence is talking about Buddhism. This falls under 290, because we're talking about another religion. However, if the sentence is "vaguely" talking about religion and doesn't fit within any of the labels, then choose 200 (Other).

\item Where do ads/exam questions fit?

In whichever of the genres you would expect to come across advertisements/exam questions. However, note that the data is scraped from the main information channel of source (i.e., advertisements next to a news text or before a vlog are not included).

\item Can we use external resources?

External resources are allowed, but do not look up the literal sentence.

\item How to pick topics (L1/L2) for fiction (genre)?

Note that the genre and topic tasks should be seen as distinct tasks. So, the genre fiction should not automatically lead to a literature topic label (unless the fiction work is about literature).

\item Some utterances seem to be taken from the same text; do we have to give them the same label, or take the contexts into account?

No, each utterance should be judged independently.
\end{itemize}

\section*{Note for L3:}
\begin{itemize}
\item For each L2, there is a finer-grained class namely L3. These numbers go in the thousands. Now, try to pick the most likely thousands' topic:
\begin{itemize}
    \item You will have to refer to the PDF (L3-1000.pdf) for the right classes.
    \item Please write the class number in the spreadsheet cell. There is no dropdown menu.
\end{itemize}
\item The "no-topic" option still exists. Use "NT";
\item You should pick the fine-grained L3 topic that best fits the utterance. This time you don’t have to match the L1-L2 categories, but we ask you to NOT update your previous L1-L2 annotations, and just annotate L3 independently.
\end{itemize}

\section{Annotation Tool}
\begin{figure*}
    \includegraphics[width=\textwidth]{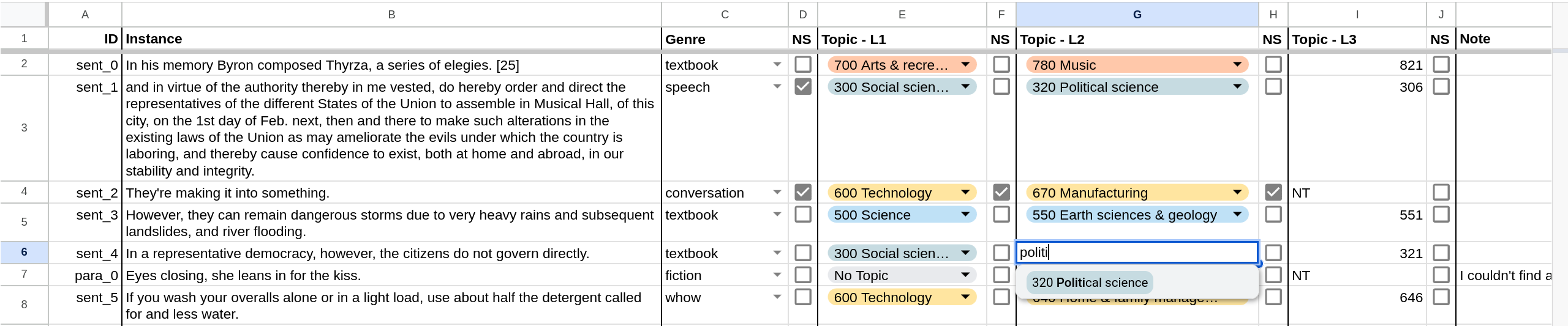}
    \caption{Example of annotation in Google Spreadsheets. NS = Not Sure}
    \label{fig:annTool}
\end{figure*}

We used Google Spreadheets for annotation. The setup is shown in Figure~\ref{fig:annTool}.

\end{document}